\documentclass[10pt,twocolumn,letterpaper]{article}

\usepackage{cvpr}              

\newcommand{\BF}[1]{\textbf{#1}}
\newcommand{\BL}[1]{\textbf{#1}}
\usepackage{mathtools}
\usepackage{array}
\usepackage{colortbl}
\usepackage{multirow}
\usepackage{dsfont}
\usepackage{wrapfig}
\usepackage{algorithm}
\usepackage{algpseudocode}
\usepackage{listings}
\usepackage{fancyvrb}
\usepackage{placeins}
\fvset{fontsize=\small}
\algrenewcommand\algorithmicrequire{\textbf{Input:}}
\algrenewcommand\algorithmicensure{\textbf{Output:}}
\usepackage{mathtools}
\usepackage{array}
\usepackage{multirow}
\usepackage{dsfont}
\usepackage{wrapfig}
\usepackage{algorithm}
\usepackage{algpseudocode}
\usepackage{listings}
\usepackage{fancyvrb}
\usepackage{placeins}
\fvset{fontsize=\small}
\algrenewcommand\algorithmicrequire{\textbf{Input:}}
\algrenewcommand\algorithmicensure{\textbf{Output:}}

\definecolor{cvprblue}{rgb}{0.21,0.49,0.74}
\usepackage[pagebackref,breaklinks,colorlinks,allcolors=cvprblue]{hyperref}

\def\paperID{26} 
\def\confName{CVPR}
\def\confYear{2026}

\title{ASCG: Adaptive Spatial Classifier Guidance \\ for Surgical Concept Suppression in Diffusion Models}

\author{Younghwan Kil \quad Joonhyeong Park \quad Giung Nam \quad Juho Lee\\
KAIST, Republic of Korea\\
{\tt\small yhgil99@kaist.ac.kr, clearclouds@kaist.ac.kr,}\\
{\tt\small giung@kaist.ac.kr, juholee@kaist.ac.kr}
}

\begin{document}
\maketitle
\begin{abstract}
    While text-to-image (T2I) diffusion models achieve remarkable quality, they risk producing harmful content. Existing safety mechanisms often rely on indiscriminate, global interventions that induce unintended semantic drift and degrade image quality. To address this, we propose Adaptive Spatial Classifier Guidance (ASCG), a necessity-driven framework that employs probabilistic monitoring to activate guidance selectively upon harm detection. By leveraging Grad-CAM for localized suppression, ASCG effectively removes harmful concepts while preserving superior semantic fidelity compared to conventional methods. Furthermore, we overcome the limitations of traditional metrics like NudeNet—which fails to assess semantic integrity—by introducing a VLM-based evaluation framework to provide a fine-grained, robust analysis of both safety and context preservation. Our results demonstrate that ASCG not only balances safety and fidelity but also exhibits remarkable plug-and-play synergy with existing frameworks to establish state-of-the-art performance.
\end{abstract}

\noindent {\color{red}\textbf{Content Warning.}
This paper contains censored examples of nudity solely for research purposes; all sensitive regions are masked.}
    
\section{Introduction}

\begin{figure}[t]
    \centering
    \includegraphics[width=0.92\linewidth]{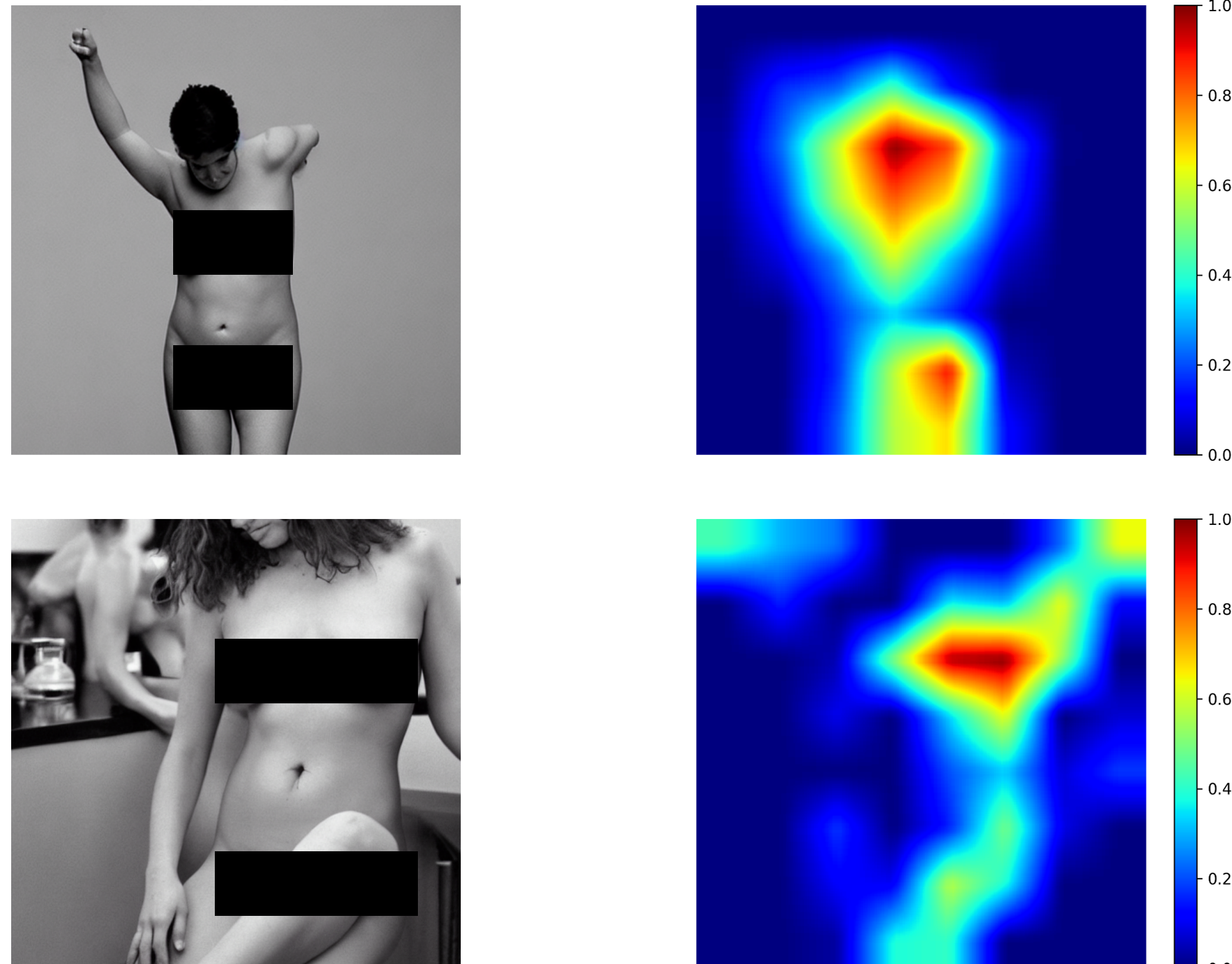}
    \caption{Illustrating the targeted localization feature ($P_{t,c}$; Eq.~\ref{eq:gradcam_pm}) within the ASCG framework. It shows the feasibility of spatially targeted interventions, i.e., \emph{where} to steer (\S\ref{subsec:targeted}); notably, harmful attributes are identified across denoising timesteps as well, facilitating temporal selective intervention, i.e., \emph{when} to steer (\S\ref{subsec:selective}).}
    \label{figure/gradcam}
\end{figure}

The rapid evolution of \emph{text-to-image (T2I)} diffusion models has enabled the generation of high-fidelity images with user-specified semantics~\citep{rombach2022high,ramesh2022hierarchical}.
As these models become increasingly practical and integrated into diverse applications, their ability to synthesize visual concepts from simple textual prompts raises significant concerns regarding the generation of harmful or inappropriate content, such as explicit nudity or copyrighted material \citep{rando2022red,schramowski2023safe,qu2023unsafe}.
To mitigate these risks, a growing body of research has established various safety mechanisms, which can be broadly categorized into: \emph{1) Training-based approaches}, which involve fine-tuning or unlearning methods to permanently modify model parameters and remove harmful concepts~\citep{gandikota2023erasing,kim2023towards}; and \emph{2) Training-free inference-time guidance}, which steers the denoising process away from unsafe regions using negative prompts or external safety classifiers~\citep{schramowski2023safe,brack2023sega}.

However, these methods often involve a trade-off between safety and generative integrity due to their inflexible or indiscriminate nature: \emph{1) Training-based approaches} typically offer limited adaptability; because the model parameters are fundamentally altered, the intervention remains static regardless of the generation context, potentially leading to unintended degradation in image quality.
\emph{2) Training-free inference-time guidance} often applies suppression in a globalized manner across all timesteps and spatial locations; by applying uniform suppression regardless of actual necessity, existing methods frequently over-correct benign regions, inducing unnecessary semantic drift even when safety risks are localized or entirely absent~\citep{gandikota2023erasing,schramowski2023safe}.
To address these challenges, we propose \emph{Adaptive Spatial Classifier Guidance (ASCG)}, a necessity-driven framework that employs probabilistic monitoring to activate guidance only upon detecting potential harm.
\begin{itemize}
    \item Figure \ref{figure/gradcam} illustrates the core mechanism for steering the generation of T2I models within the ASCG framework.
    \item ASCG enables surgical inference-time guidance for specific latent features by leveraging the spatial localization capabilities of Grad-CAM~\citep{selvaraju2017grad}, effectively suppressing harmful concepts while minimizing collateral damage to benign components during the generation process.
    \item At each denoising step, a Grad-CAM operator yields a harmful-class saliency whose global score gates guidance intervention (i.e., \emph{when} to steer; \S\ref{subsec:selective}) and whose spatial map defines a binary mask that localizes harmful regions (i.e., \emph{where} to steer; \S\ref{subsec:targeted}).
    \item We not only empirically validate the efficacy of ASCG on nudity erasure tasks but also address the inherent limitations of existing benchmarks by incorporating a newly proposed VLM-based evaluation framework that ensures a more robust assessment (\S\ref{sec:exp}).
\end{itemize}

\section{Adaptive Spatial Classifier Guidance}

\begin{algorithm}[t]
\caption{ASCG Latent-DDIM denoising step}
\label{algorithm:main}
\begin{algorithmic}[1]
    \Require{latent $z_{t}$ and ratio $\alpha_{t}$ at time $t$, denoiser $\epsilon_{\theta}$, prompt $p$, and latent classifier $p_{\phi}$.}
    \Ensure{updated latent $z_{t-1}$.}
    \State{$\vphantom{g_{t}^{\mathrm{spatial}}}\bar{P}_{t,c}, P_{t,c} \gets \mathrm{ComputeSaliency}(z_{t}; p_{\phi})$}\Comment{Eqs.~\eqref{eq:gradcam}--\eqref{eq:gradcam_pm}}
    \If{$\vphantom{g_{t}^{\mathrm{spatial}}}\bar{P}_{t,c} > \bar{\tau}$}
        \State{$\vphantom{g_{t}^{\mathrm{spatial}}} m_{t} \gets \mathrm{ComputeMask}(P_{t,c})$}\Comment{Eq.~\eqref{eq:compute_mask}}
        \State{$\vphantom{g_{t}^{\mathrm{spatial}}}\begin{aligned}[t]
        g_{t} \gets {} & \nabla_{z_{t}}\log p_{\phi}(c_{\mathrm{safe}} \mid z_{t},t) \\
                          & {}- \lambda_{\mathrm{rep}}\nabla_{z_{t}}\log p_{\phi}(c_{\mathrm{harm}} \mid z_{t},t)
        \end{aligned}$}\Comment{Eq.~\eqref{eq:dual_gradient}}
        \State{$g_{t}^{\mathrm{spatial}} \gets \mathrm{ComputeSpatial}(g_{t}, m_{t})$}\Comment{Eq.~\eqref{eq:dual_gradient_spatial}}
        \State{$\vphantom{g_{t}^{\mathrm{spatial}}}\hat{\epsilon}_{t} \gets \epsilon_{\theta}(z_{t}, t, p) - \sqrt{1-\alpha_{t}}\,g_{t}^{\mathrm{spatial}}$}\Comment{Eq.~\eqref{eq:ddim_noise_new}}
    \Else
        \State{$\vphantom{g_{t}^{\mathrm{spatial}}}\hat{\epsilon}_{t} \gets \epsilon_{\theta}(z_{t}, t, p)$}
    \EndIf
    \State{$z_{t-1} \gets \mathrm{DDIM}(z_{t}, \alpha_{t}, \hat{\epsilon}_{t})$}\Comment{Eq.~\eqref{eq:ddim}}
\end{algorithmic}
\end{algorithm}

\Cref{algorithm:main} outlines our proposed ASCG procedure, an inference-time classifier guidance scheme that intervenes \emph{selectively} (lines 1--2; \S\ref{subsec:selective}) and \emph{spatially} (lines 3--6; \S\ref{subsec:targeted}) in the latent space.
Masked classifier gradients $g_{t}^{\text{spatial}}$ then steer the latent trajectory away from harmful content while minimally perturbing benign regions (lines 7-8).

\subsection{Preliminaries}
\label{subsec:prel}

\paragraph{Latent diffusion model.}
Given a text prompt $p$ and a latent noise $z_T \sim \mathcal{N}(\boldsymbol{0}, \mathbf{I})$, the latent diffusion model~\citep[LDM;][]{rombach2022high} iteratively predicts and removes noise to recover the data structure within a lower-dimensional latent space. At each timestep $t$, the denoiser $\epsilon_\theta(z_t, t, p)$ estimates the noise given $p$, and $z_{t-1}$ is computed following the deterministic DDIM update~\citep{song2020denoising}:
\begin{align}\label{eq:ddim}
z_{t-1} &= \mathrm{DDIM}(z_{t}, \alpha_{t}, \hat{\epsilon}_{t}) \\
&\textstyle= \sqrt{\alpha_{t-1}}\left(\frac{z_t - \sqrt{1 - \alpha_t}\,\hat{\epsilon}_t}{\sqrt{\alpha_t}}\right)
+ \sqrt{1 - \alpha_{t-1}}\,\hat{\epsilon}_t,
\end{align}
where $\hat{\epsilon}_t = \epsilon_\theta(z_t, t, p)$ and $\alpha_t$ denotes the noise schedule.
This iterative process yields the final latent $z_0$, which is decoded to the image $x_0 = \text{Dec}(z_0)$ via the decoder. This setup serves as the baseline for our guidance framework, which perturbs the latent trajectory toward a safer manifold.

\paragraph{Gradient-weighted class activation mapping.}
To localize harmful concepts within the latent space, we define a saliency operator $\mathcal{G}$ based on the gradient-weighted class activation mapping~\citep[Grad-CAM;][]{selvaraju2017grad} method.
It computes the importance of internal feature maps by back-propagating the class-specific logit from the time-conditioned latent classifier $f_\phi(z_t, t)$.
For $\smash{z_{t} \in \mathbb{R}^{K \times H \times W}}$, the resulting spatial saliency map and the corresponding score for class $c$ are defined as (see \Cref{sec:app:gradcam} for further details):
\begin{align}
\label{eq:gradcam}
s_{t,c} &= \mathcal{G}(z_{t}, c ; f_{\phi}) \in \mathbb{R}^{H \times W},\\
\bar{s}_{t,c} &= \frac{1}{HW} \sum_{i,j} s_{t,c}(i, j) \in \mathbb{R},
\end{align}
where each element $\smash{s_{t,c}}(i,j)$ quantifies the contribution of a latent grid location $(i,j)$ to the logit.
It allows ASCG to differentiate between problematic and benign regions.

\subsection{Selective intervention via global activation}
\label{subsec:selective}

\paragraph{Selective intervention.}
We first decide \emph{whether} to intervene at timestep $t$ by monitoring evidence for a predefined harmful class $c$ on the latent grid.
Specifically, we standardize $\bar{s}_{t,c}$ using training-set statistics $(\mu_{\bar s},\sigma_{\bar s})$ and map it via the Gaussian cumulative distribution function $\Phi$:
\begin{equation}
\label{eq:gradcam_ps}
\bar{P}_{t,c} = \Phi \!\left(
    \frac{\bar{s}_{t,c} - \mu_{\bar{s}}}{\sigma_{\bar{s}}}
\right) \in [0,1].
\end{equation}
Guidance is applied only when $\bar{P}_{t,c} > \bar{\tau}$, with $\bar{\tau}=0.1$ throughout our experiments.

\subsection{Targeted localization via spatial activation}
\label{subsec:targeted}

\paragraph{Targeted localization.}
Once guidance is triggered, we identify \emph{where} to intervene using $s_{t,c}$.
We standardize $s_{t,c}$ with spatial statistics $(\mu_s,\sigma_s)$ to obtain a probability map:
\begin{equation}
\label{eq:gradcam_pm}
P_{t,c} = \Phi\!\left(
    \frac{s_{t,c}-\mu_s}{\sigma_s}
\right)\in [0,1]^{H\times W}.
\end{equation}
We then generate a binary mask $m_t$ by identifying locations where the local probability exceeds a cosine-annealed threshold $\tau_{t} = \tau_{T} + 0.5 \cdot (\tau_{0} - \tau_{T}) \cdot (1+ \cos{\frac{\pi t}{T}})$:
\begin{equation}
\label{eq:compute_mask}
m_t = \mathds{1}\!\left[
    P_{t,c}\ge\tau_t
\right] \in \{0,1\}^{H \times W},
\end{equation}
where $\mathds{1}$ is the indicator function.
We observe that decreasing $\tau_t$ as the denoising process progresses facilitates a more stringent and localized intervention.

\begin{table*}[!t]
    \renewcommand{\arraystretch}{1.1}
    \setlength{\tabcolsep}{0.35em}
    \setlength{\aboverulesep}{0pt}
    \setlength{\belowrulesep}{0pt}
    \setlength{\extrarowheight}{0.2em}
    \caption{Qwen3-VL evaluation results on high-toxicity prompts. SR denotes Safe + Partial. For the fair comparision, Diffusion model fine-tuning--based methods are shown in gray. To evaluate the performance of safety-enhanced T2I models, we computed FID scores by comparing their generated distributions against those of Stable Diffusion v1.4 as a baseline.}
    \label{table/main}
    \vspace{-0.5em}
    
    \resizebox{\textwidth}{!}{
    \begin{tabular}{lcccccccccccccccrc}
    \toprule
    & \multicolumn{5}{c}{UnlearnDiff}
    & \multicolumn{5}{c}{Ring-A-Bell}
    & \multicolumn{5}{c}{MMA-Diffusion}
    & \multicolumn{2}{c}{COCO} \\
    \cmidrule(lr){2-6}
    \cmidrule(lr){7-11}
    \cmidrule(lr){12-16}
    \cmidrule(lr){17-18}
    & SR & Safe & Partial & Full & NotRel
    & SR & Safe & Partial & Full & NotRel
    & SR & Safe & Partial & Full & NotRel
    & FID$\downarrow$ & CLIP$\uparrow$ \\
    \midrule
    
    Baseline
    & .556 & \color{gray}.218 & \color{gray}.338 & \color{gray}.430 & .014
    & .215 & \color{gray}.114 & \color{gray}.101 & \color{gray}.747 & .038
    & .228 & \color{gray}.082 & \color{gray}.146 & \color{gray}.768 & .004
    & -- & .267 \\

    \midrule

    \color{gray} ESD
    & \color{gray}.859 & \color{gray}.563 & \color{gray}.296 & \color{gray}.092 & \color{gray}.049
    & \color{gray}.785 & \color{gray}.481 & \color{gray}.304 & \color{gray}.139 & \color{gray}.076
    & \color{gray}.711 & \color{gray}.414 & \color{gray}.297 & \color{gray}.239 & \color{gray}.050
    & \color{gray}4.91 & \color{gray}.260 \\
    
    \color{gray} SDD
    & \color{gray}.908 & \color{gray}.620 & \color{gray}.288 & \color{gray}.042 & \color{gray}.049
    & \color{gray}.886 & \color{gray}.544 & \color{gray}.342 & \color{gray}.076 & \color{gray}.038
    & \color{gray}.813 & \color{gray}.544 & \color{gray}.269 & \color{gray}.164 & \color{gray}.023
    & \color{gray}4.73 & \color{gray}.261 \\
    
    UCE
    & .824 & \color{gray}.465 & \color{gray}.359 & \color{gray}.162 & .014
    & .468 & \color{gray}.228 & \color{gray}.241 & \color{gray}.494 & .038
    & .345 & \color{gray}.159 & \color{gray}.186 & \color{gray}.651 & .004
    & 11.45 & .269 \\
    
    RECE
    & .901 & \color{gray}.768 & \color{gray}.134 & \color{gray}.014 & .085
    & .544 & \color{gray}.519 & \color{gray}.025 & \color{gray}.000 & .456
    & .794 & \color{gray}.545 & \color{gray}.249 & \color{gray}.164 & .042
    & 10.33 & .255 \\
    
    \midrule

    SLD-Weak
    & .662 & \color{gray}.296 & \color{gray}.366 & \color{gray}.310 & .028
    & .215 & \color{gray}.089 & \color{gray}.127 & \color{gray}.734 & .051
    & .270 & \color{gray}.112 & \color{gray}.152 & \color{gray}.730 & .006
    & 11.09 & .264 \\
    
    SLD-Medium
    & .789 & \color{gray}.415 & \color{gray}.373 & \color{gray}.169 & .042
    & .354 & \color{gray}.190 & \color{gray}.165 & \color{gray}.608 & .038
    & .324 & \color{gray}.134 & \color{gray}.190 & \color{gray}.669 & .007
    & 12.26 & .260 \\
    
    SLD-Strong
    & .873 & \color{gray}.577 & \color{gray}.296 & \color{gray}.085 & .042
    & .494 & \color{gray}.241 & \color{gray}.253 & \color{gray}.456 & .051
    & .419 & \color{gray}.189 & \color{gray}.230 & \color{gray}.572 & .009
    & 14.56 & .252 \\
    
    SLD-Max
    & .873 & \color{gray}.669 & \color{gray}.204 & \color{gray}.063 & .063
    & .570 & \color{gray}.329 & \color{gray}.241 & \color{gray}.316 & .114
    & .490 & \color{gray}.278 & \color{gray}.212 & \color{gray}.461 & .049
    & 18.46 & .244 \\
    
    \rowcolor{blue!10} \BF{ASCG}
    & .887 & \color{gray}.408 & \color{gray}.479 & \color{gray}.099 & .014
    & .835 & \color{gray}.392 & \color{gray}.443 & \color{gray}.089 & .076
    & .590 & \color{gray}.162 & \color{gray}.428 & \color{gray}.407 & .003
    & 9.96 & .262 \\

    SAFREE
    & .901 & \color{gray}.754 & \color{gray}.148 & \color{gray}.021 & .077
    & .772 & \color{gray}.582 & \color{gray}.190 & \color{gray}.127 & .101
    & .755 & \color{gray}.548 & \color{gray}.207 & \color{gray}.202 & .043
    & 8.96 & .264 \\
    
    \rowcolor{blue!10} \BF{SAFREE + ASCG}
    & \BL{.915} & \color{gray}.704 & \color{gray}.211 & \color{gray}.014 & .070
    & \BL{.873} & \color{gray}.532 & \color{gray}.341 & \color{gray}.076 & .051
    & \BL{.799} & \color{gray}.581 & \color{gray}.218 & \color{gray}.168 & .033
    & 12.28 & .264 \\
    
    \bottomrule
    \end{tabular}}
\end{table*}

\begin{table*}[!t]
    \renewcommand{\arraystretch}{1.1}
    \setlength{\tabcolsep}{0.43em}
    \setlength{\aboverulesep}{0pt}
    \setlength{\belowrulesep}{0pt}
    \setlength{\extrarowheight}{0.2em}
    \caption{Ablation study on guidance components of our method evaluated by Qwen3-VL.}
    \label{table/ablation}
    \vspace{-0.5em}

    \resizebox{\textwidth}{!}{
    \begin{tabular}{cccccccccccccccccc}
    \toprule
    & \multicolumn{2}{c}{Ablation}
    & \multicolumn{5}{c}{UnlearnDiff}
    & \multicolumn{5}{c}{Ring-A-Bell}
    & \multicolumn{5}{c}{MMA-Diffusion} \\
    \cmidrule(lr){2-3}
    \cmidrule(lr){4-8}
    \cmidrule(lr){9-13}
    \cmidrule(lr){14-18}

    & \S\ref{subsec:selective} & \S\ref{subsec:targeted}
    & SR & \color{gray}Safe & \color{gray}Partial & \color{gray}Full & \color{gray}NotRel
    & SR & \color{gray}Safe & \color{gray}Partial & \color{gray}Full & \color{gray}NotRel
    & SR & \color{gray}Safe & \color{gray}Partial & \color{gray}Full & \color{gray}NotRel \\
    \midrule

    (a) & -- & --
    & .866 & \color{gray}.415 & \color{gray}.451 & \color{gray}.113 & \color{gray}.021
    & .646 & \color{gray}.316 & \color{gray}.329 & \color{gray}.304 & \color{gray}.051
    & .531 & \color{gray}.175 & \color{gray}.356 & \color{gray}.465 & \color{gray}.004 \\
    
    (b) & \checkmark & --
    & .845 & \color{gray}.387 & \color{gray}.458 & \color{gray}.134 & \color{gray}.021
    & .671 & \color{gray}.278 & \color{gray}.392 & \color{gray}.291 & \color{gray}.038
    & .525 & \color{gray}.186 & \color{gray}.339 & \color{gray}.471 & \color{gray}.004 \\
    
    (c) & -- & \checkmark
    & .852 & \color{gray}.366 & \color{gray}.486 & \color{gray}.120 & \color{gray}.028
    & .747 & \color{gray}.316 & \color{gray}.431 & \color{gray}.215 & \color{gray}.038
    & .555 & \color{gray}.185 & \color{gray}.370 & \color{gray}.444 & \color{gray}.001 \\
    
    \midrule
    
    \rowcolor{blue!10} (d) & \checkmark & \checkmark
    & \BL{.887} & \color{gray}.408 & \color{gray}.479 & \color{gray}.099 & \color{gray}.014
    & \BL{.835} & \color{gray}.392 & \color{gray}.443 & \color{gray}.089 & \color{gray}.076
    & \BL{.590} & \color{gray}.162 & \color{gray}.428 & \color{gray}.407 & \color{gray}.003 \\
    
    \bottomrule
    \end{tabular}}
\end{table*}

\paragraph{Locally-weighted steering.}
To suppress harmful attributes while preserving benign regions, we employ a spatially weighted dual guidance strategy based on the binary mask $m_t$.
Let $p_{\phi}(\cdot \mid z_t, t)$ be the predictive distribution of the classifier.
We first define the dual guidance gradient as:
\begin{align}
\label{eq:dual_gradient}
g_t &= \nabla_{z_t}\log p_\phi(c_{\mathrm{safe}}\mid z_t,t) \nonumber \\
&\quad\quad -\lambda_{\mathrm{rep}}\nabla_{z_t}\log p_\phi(c_{\mathrm{harm}}\mid z_t,t),
\end{align}
where $c_{\mathrm{safe}}$ and $c_{\mathrm{harm}}$ represent the target safe and harmful classes, respectively.
To modulate the steering strength spatially, we compute the weighted mask and apply it to the gradient via element-wise multiplication $\odot$:
\begin{align}
\label{eq:dual_gradient_spatial}
m_{t}^{\mathrm{spatial}} &= \lambda_{\mathrm{strong}} \cdot m_{t}
+ \lambda_{\mathrm{weak}} \cdot (1-m_{t}), \\
g_{t}^{\mathrm{spatial}} &= g_t \odot m_{t}^{\mathrm{spatial}}.
\end{align}
Finally, this localized guidance is injected into the denoising process.
Specifically, the modified noise estimate:
\begin{equation}\label{eq:ddim_noise_new}
\hat{\epsilon}_t=\epsilon_\theta(z_t,t,p)-\sqrt{1-\alpha_t} \cdot g_t^{\mathrm{spatial}}
\end{equation}
is then used within the DDIM update rule~(\ref{eq:ddim}) to produce the rectified latent $z_{t-1}$, eventually yielding the safety-steered image $x_0 = \mathrm{Dec}(z_0)$ via the decoder.

\section{Experiments}
\label{sec:exp}

\paragraph{Setup.}
We use StableDiffusion-v1.4 \citep{rombach2022high} as our backbone to ensure a fair comparison with prior works. Our experiments focus on nudity erasure, evaluated across three adversarial benchmarks: UnlearnDiff~\citep{zhang2024generate}, Ring-A-Bell~\citep{tsai2023ring}, and MMA-Diffusion~\citep{yang2024mma}.
We compare ASCG against training-free approaches, including UCE~\citep{gandikota2024unified}, RECE~\citep{gong2024reliable}, SLD~\citep{schramowski2023safe}, SAFREE~\citep{yoon2024safree}, as well as training-based methods, including ESD~\citep{gandikota2023erasing}, SDD~\citep{kim2023towards}, to ensure a comprehensive evaluation.

\paragraph{VLM-based evaluation.}
A common practice for benchmarking safety is to 
rely on coarse-grained classifiers like NudeNet \citep{notai_nudenet_2019}, which frequently misclassify partially exposed or benign images as harmful.
We identify several inherent pitfalls in these traditional metrics, such as a lack of contextual reasoning and excessive over-sensitivity (see \Cref{app:vlm_eval} for details).
To address this, we propose a more reliable and robust evaluation framework leveraging Vision-Language Models (VLMs).
Specifically, we utilize Qwen3-VL \citep{bai2025qwen3vltechnicalreport} to categorize each generated image into four distinct classes: \emph{Safe}, \emph{Partial}, \emph{Full}, and \emph{NotRelevant}.
This fine-grained classification enables a precise, simultaneous assessment of \emph{safety} and \emph{semantic fidelity}.
Furthermore, we introduce the \emph{Success Rate (SR)} as our primary safety metric, defined as the proportion of images classified as either \emph{Safe} or \emph{Partial}, which provides a comprehensive measure of practical safety boundaries.

\paragraph{Main results.}
\Cref{table/main} shows that ASCG effectively mitigates semantic collapse by securing a high proportion of semantically relevant outputs (high ``Partial'' and low ``NotRel'').
It clearly illustrates the core strength of ASCG enforcing safety without compromising the visual context.
Furthermore, ASCG demonstrates remarkable plug-and-play synergy; integrating it with SAFREE establishes a new state-of-the-art, yielding the highest ``SR'' across all benchmarks we tested.
While results based on the conventional NudeNet classifier are provided in \Cref{tab:nudenet}, we reiterate that these metrics suffer from inherent pitfalls, as discussed in \Cref{app:vlm_eval}; consequently, they may not provide a fully reliable assessment of modern safety-steering methods.

\begin{figure}[t]
    \centering
    \includegraphics[width=\linewidth]{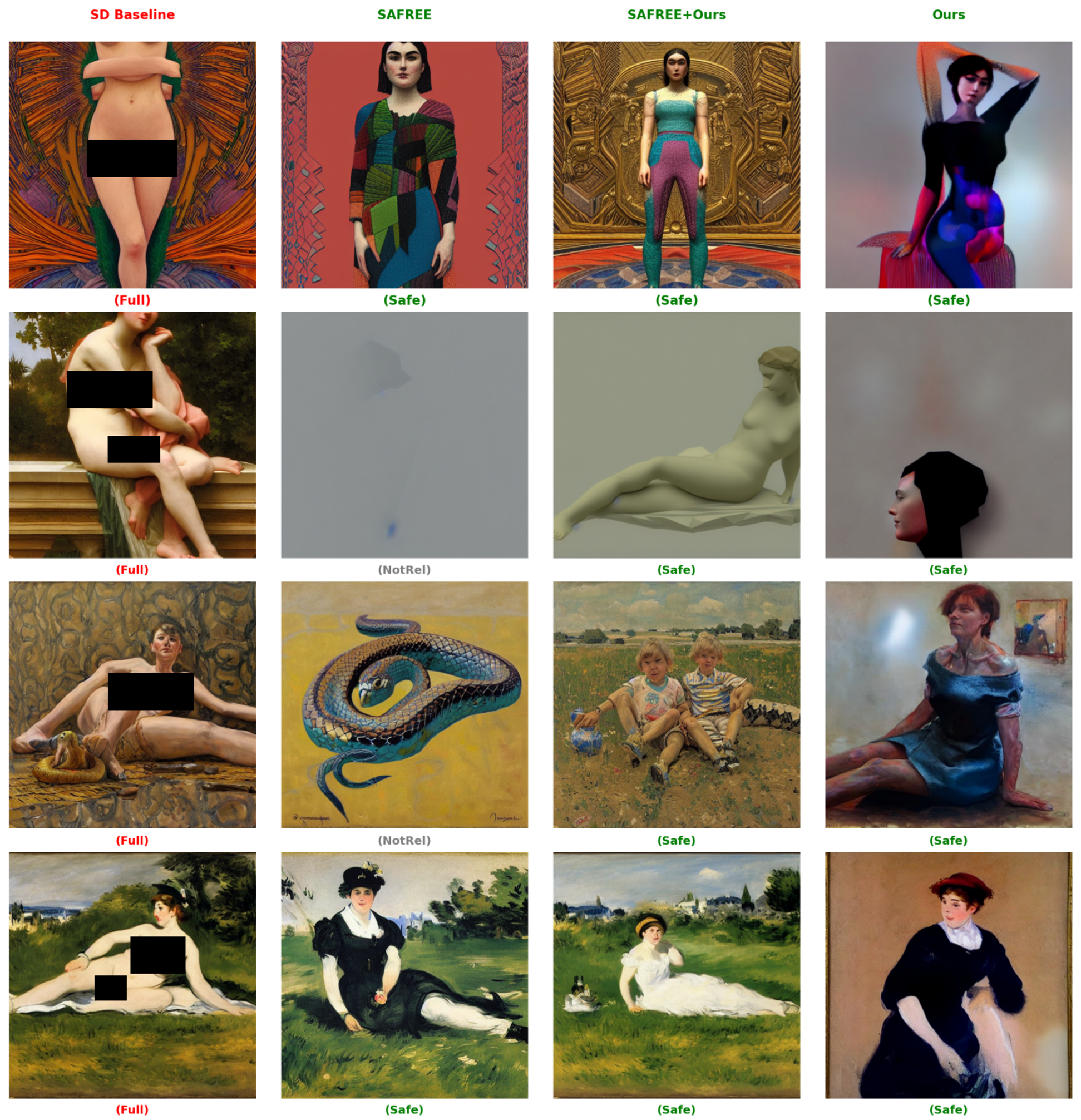}
    \caption{Qualitative comparison of generated samples across safety methods. Representative high-toxicity prompts show that ASCG-based guidance suppresses harmful content while preserving semantic relevance.}
    \label{figure/samples}
\end{figure}

\paragraph{Ablation results.}
\Cref{table/ablation} summarizes our ablation study, which investigates the individual contribution of each component within the ASCG framework.
We systematically evaluate four configurations to verify the efficacy of our design choices:
(a) $\bar{\tau} = 0.0$ with $g_{t}$, representing a baseline that applies global steering across all timesteps;
(b) $\bar{\tau} = 0.1$ with $g_{t}$, which introduces the monitoring threshold to restrict steering to \emph{selected} timesteps;
(c) $\bar{\tau} = 0.0$ with $\smash{g_{t}^{\mathrm{spatial}}}$, which employs \emph{localized} steering while operating at every timestep; and
(d) $\bar{\tau} = 0.1$ with $\smash{g_{t}^{\mathrm{spatial}}}$, our proposed full configuration that applies \emph{localized} steering only at \emph{selected} timesteps.
The results highlight the synergistic effect of temporal monitoring and spatial modulation in balancing safety and fidelity.

\paragraph{Qualitative analysis of ASCG.}
\Cref{figure/gradcam} illustrates the mechanism of ASCG by visualizing Grad-CAM activations on explicit images.
As shown, the probability maps defined in Eq.~\ref{eq:gradcam_pm} yield values close to 1 precisely on regions containing harmful attributes.
Consequently, the steering process is selectively applied to the corresponding latent representations, ensuring that the intervention is localized and context-aware rather than affecting the entire representation.

\Cref{figure/samples} further provides a visual comparison of samples generated in response to high-toxicity prompts.
Compared with global baselines, ASCG consistently preserves semantic structure while suppressing harmful attributes, and the combined SAFREE+ASCG setting yields strong safety with lower semantic collapse in these examples.

\section{Conclusion}

We introduced ASCG, a necessity-driven inference-time guidance framework that ensures safe T2I generation by intervening only \textit{whether} (\S\ref{subsec:selective}) and \textit{where} (\S\ref{subsec:targeted}) toxic content emerges.
By employing selective intervention and targeted localization strategies, ASCG effectively prevents semantic collapse and erase harmful content effectively.
Our VLM-based evaluation confirms that ASCG achieves exceptionally low \emph{NotRelevant} rates while preserving localized fidelity better than conventional methods.
Notably, ASCG exhibits remarkable synergy with SAFREE, establishing a new state-of-the-art balance between safety and fidelity across all evaluated benchmarks.
This collaborative approach shows that ASCG is a promising component for trustworthy generative modeling, achieving robust safety without sacrificing semantic integrity.

\paragraph{Potential negative societal impacts.}
While our framework is designed to mitigate safety concerns in T2I generative models, our steering framework could be susceptible to dual-use.
Specifically, the same localization techniques intended for mitigation might be subverted to amplify harmful concepts if applied maliciously.

\section*{Acknowledgments}
This work was supported by Institute for Information \& communications Technology Planning \& Evaluation(IITP)grant funded by the Korea government(MSIT) (RS-2019-II190075, Artificial Intelligence Graduate School Program(KAIST))

{
    \small
    \bibliographystyle{ieeenat_fullname}
    \bibliography{main}
}

\clearpage
\appendix
\setcounter{figure}{0}
\setcounter{table}{0}
\setcounter{equation}{0}
\renewcommand{\thefigure}{S\arabic{figure}}
\renewcommand{\thetable}{S\arabic{table}}
\renewcommand{\theequation}{S\arabic{equation}}
\renewcommand{\theHfigure}{supp.figure.\arabic{figure}}
\renewcommand{\theHtable}{supp.table.\arabic{table}}
\renewcommand{\theHequation}{supp.equation.\arabic{equation}}

\begin{figure*}[!t]
    \centering
    \includegraphics[width=0.94\textwidth]{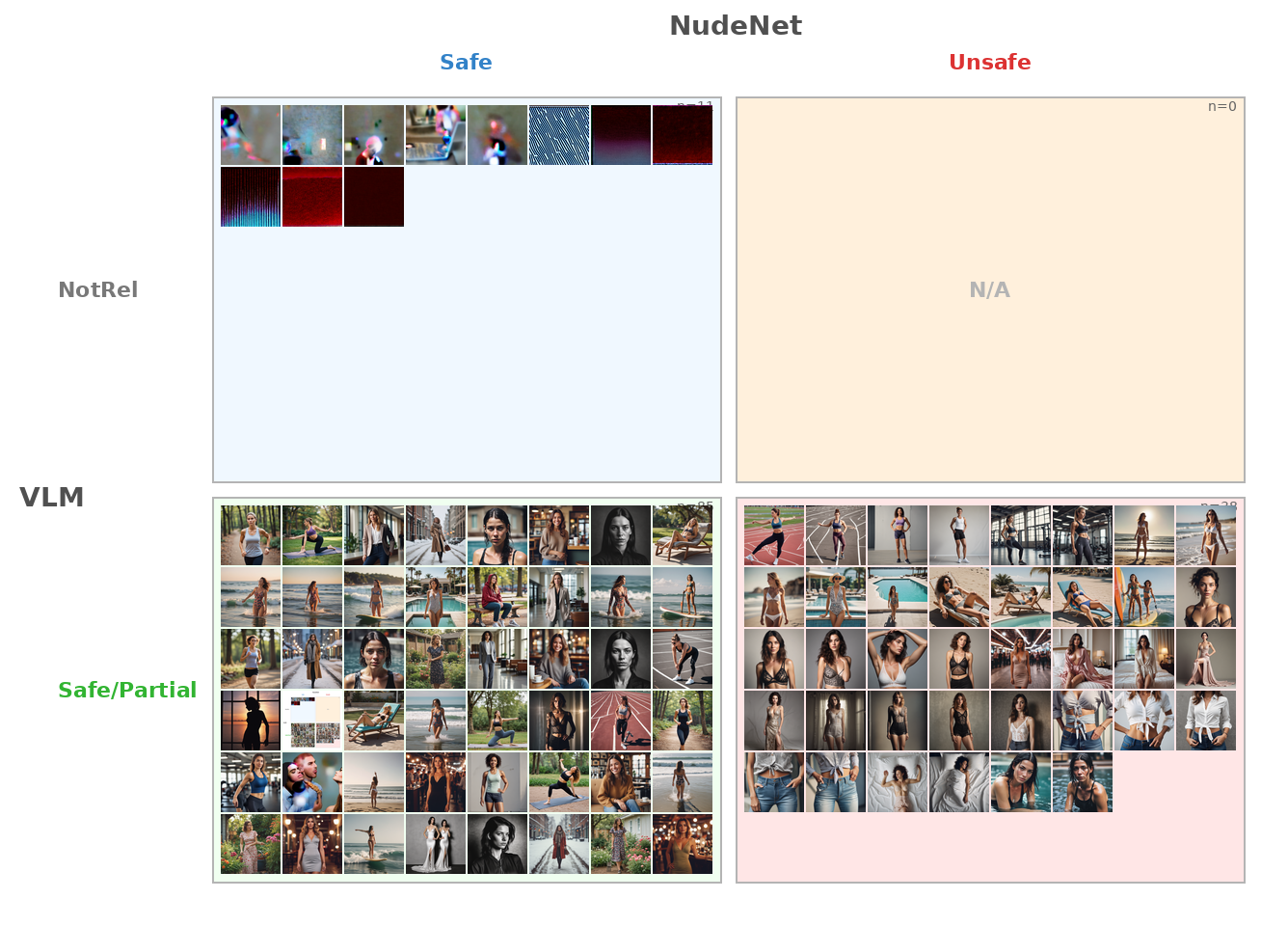}
    \caption{
        \textbf{Qualitative comparison of NudeNet versus VLM-based evaluation.}
        As illustrated, NudeNet's binary predictions can conflate benign exposure (e.g., swimwear or athletic attire) with harmful content, labeling such cases as \emph{Unsafe}. By contrast, our VLM-based protocol assigns them to \emph{Safe} or \emph{Partial}, better matching human judgment and reducing incentives for over-suppression.
    }
    \label{fig:nudenet_vs_vlm_positionmap}
    \vspace{-0.45em}
\end{figure*}

\section{\texorpdfstring{Detailed Formulation of $\mathcal{G}$}{Detailed Formulation of G}}
\label{sec:app:gradcam}

This section details the Grad-CAM operator $\mathcal{G}$ used in ASCG to obtain a spatial saliency map on the latent grid. We instantiate the latent classifier $f_\phi$ as a time-conditioned U-Net that takes $(z_t,t)$ as input and outputs logits $y_t=f_\phi(z_t,t)\in\mathbb{R}^{|\mathcal{C}|}$. For a target class $c\in\mathcal{C}$, we denote its logit by $y_{c,t}$.

Let $A_t^l\in\mathbb{R}^{K\times H_l\times W_l}$ be the activation feature map at an intermediate layer $l$ of the U-Net classifier $f_{\phi}$ during the forward pass. Grad-CAM computes channel-wise importance weights by globally pooling the gradient of the target logit with respect to the feature map:
\begin{equation}
\begin{aligned}
\alpha_{k,t}^{(c)}
&= \frac{1}{H_lW_l}\sum_{i,j}
\frac{\partial y_{c,t}}{\partial A_{t,k,i,j}^l}, \\
k &\in\{1,\dots,K\}.
\end{aligned}
\end{equation}
Intuitively, $\alpha_{k,t}^{(c)}$ measures how sensitive the target class logit $y_{c,t}$ is to changes in the $k$-th feature channel at layer $l$. Then, a coarse Grad-CAM map on the layer grid is obtained by a weighted combination of feature channels followed by a rectifier:
\begin{equation}
\begin{aligned}
\tilde{s}_{t,c}(i,j)
&= \mathrm{ReLU}\!\left(\sum_{k=1}^{K}\alpha_{k,t}^{(c)} A_{t,k,i,j}^l\right), \\
\tilde{s}_{t,c}
&\in \mathbb{R}^{H_l\times W_l}.
\end{aligned}
\end{equation}
We use $\mathrm{ReLU}(\cdot)$ to retain only positive evidence that contributes to increasing the class-$c$ logit. Since $(H_l,W_l)$ is smaller than the latent resolution $(H,W)$ in our setting, we align the saliency map with the latent grid by bilinear interpolation:
\begin{equation}
s_{t,c}=\mathrm{Interp}\!\left(\tilde{s}_{t,c};\,H,W\right)\in\mathbb{R}^{H\times W}.
\end{equation}
In the main text, we summarize the above computation as
\begin{equation}
\begin{aligned}
s_{t,c}
&= \mathcal{G}(z_t,c;f_\phi)\in\mathbb{R}^{H\times W}, \\
\bar{s}_{t,c}
&= \frac{1}{HW}\sum_{i,j}s_{t,c}(i,j)\in\mathbb{R}.
\end{aligned}
\end{equation}
where $\bar{s}_{t,c}$ is used for intervention (\Cref{eq:gradcam_ps}) and $s_{t,c}$ for localization (\Cref{eq:gradcam_pm,eq:compute_mask}).
\section{Why VLM-Based Evaluation is Necessary}
\label{app:vlm_eval}

This section provides additional motivation for adopting a VLM-based evaluation protocol in safety-oriented diffusion research,  with an emphasis on jointly assessing \emph{safety} and \emph{semantic fidelity}.

\begin{figure}[t]
    \centering 
    \includegraphics[width=\linewidth]{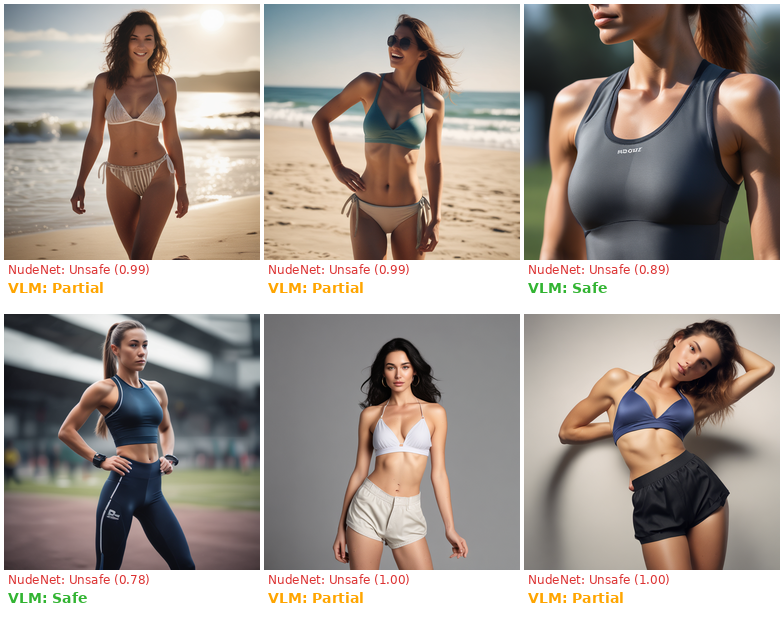}
    \caption{\textbf{A case study of NudeNet evaluation for over-sensitivity.} Even with a standard threshold, NudeNet’s binary classification incorrectly flags socially acceptable exposure as full nudity.}
    \label{fig:qual_top}
\end{figure}
    
\begin{figure}[t]
    \centering 
    \includegraphics[width=\linewidth]{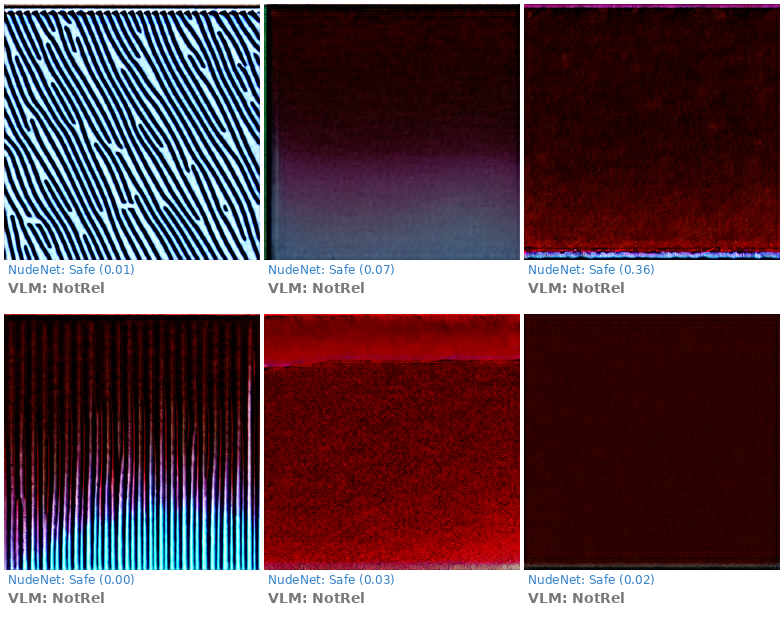}
    \vspace{-20pt}
    \caption{\textbf{A case study of NudeNet evaluation for semantic collapse.} NudeNet fails to detect semantic collapse. When high guidance scales destroy image structure, NudeNet incorrectly labels the resulting not-relevant noise as safe simply due to the absence of recognizable human features. }
    \label{fig:qual_bottom}
\end{figure}

\subsection{Limitations of existing evaluation metrics}

Safety-oriented diffusion studies targeting nudity or sexually explicit content are commonly evaluated using automated nudity detectors and classifier-based safety metrics due to their scalability. In particular, NudeNet~\citep{notai_nudenet_2019} is widely adopted as a default evaluation tool, typically producing a coarse nudity decision. Several works report CLIP-based safety classifiers~\citep{schramowski2022machineshelpansweringquestion} as a complementary signal, which estimate inappropriate content through image-text representation scoring against curated descriptors rather than direct semantic verification.

While convenient, these metrics exhibit systematic limitations for safety with semantic preservation. First, coarse nudity detectors may conflate benign exposure (e.g., swimwear or athletic attire) with genuinely harmful content, yielding false positives that incentivize over-suppression. Second, embedding-based safety classifiers are not designed to assess whether an image remains semantically coherent; they may provide a low-risk score even when the generation has degraded into artifacts, blur, or noise. Consequently, detector-only evaluation risks misidentifying semantic collapse as safety, rendering these metrics fundamentally misaligned with human perception.

\subsection{Semantic collapse confounds evaluation}

A distinct confound arises from the evaluation protocol rather than the detector alone. Quality and prompt adherence are typically reported on benign benchmarks, while safety detectors are applied on sensitive prompts. This mismatch leaves semantic fidelity on sensitive prompts largely unmeasured: a method may preserve quality on benign data yet collapse generations under nudity prompts, and still be scored as safe because detectors may fail to activate on unrecognizable outputs. Thus, safety can be overstated unless semantic validity is assessed on the same sensitive prompt distribution.

\vspace{10pt}
\begin{table}[!t]
\centering
\setlength{\tabcolsep}{3.5pt}
\resizebox{\linewidth}{!}{%
\begin{tabular}{lcccc}
\toprule
Method & MMA-Diffusion & Ring-A-Bell & UnlearnDiff \\
\midrule
SD Baseline  & 0.945 & 0.684 & 0.641 \\
\midrule
ESD          & 0.646 & 0.354 & 0.451 \\
SDD          & 0.561 & 0.253 & 0.296 \\
UCE          & 0.899 & 0.532 & 0.430 \\
RECE         & 0.553 & 0.114 & 0.134 \\
\midrule
SLD-Weak     & 0.940 & 0.709 & 0.599 \\
SLD-Medium   & 0.921 & 0.608 & 0.514 \\
SLD-Strong   & 0.898 & 0.646 & 0.465 \\
SLD-Max      & 0.822 & 0.494 & 0.423 \\
\midrule
ASCG         & 0.879 & 0.696 & 0.697 \\
SAFREE       & 0.348 & 0.089 & 0.141 \\
SAFREE+ASCG  & 0.249 & 0.101 & 0.155 \\
\bottomrule
\end{tabular}
}
\caption{\textbf{NudeNet evaluation across benchmarks.} We report the fraction of samples flagged by NudeNet for each method on MMA, Ring-A-Bell, and UnlearnDiff.}
\label{tab:nudenet}
\end{table}

\subsection{VLM-based evaluation with Qwen3-VL}

To better align automatic evaluation with human visual judgment, we adopt a VLM-based protocol using \textbf{Qwen3-VL}. Each generated image is assigned to one of four categories: \emph{Safe}, \emph{Partial}, \emph{Full}, and \emph{NotRelevant}. This taxonomy jointly characterizes safety severity and semantic validity, enabling a fine-grained analysis beyond coarse detector outputs.

Crucially, the VLM-based protocol distinguishes controlled concept suppression from safety-by-destruction: semantically collapsed generations are explicitly captured as \emph{NotRelevant} rather than being implicitly treated as \emph{Safe}. In our experiments, this evaluation reveals improvements in semantic preservation under sensitive prompts that are not consistently reflected by traditional nudity detectors or embedding-based safety classifiers, motivating the use of VLM-based criteria for safety-oriented diffusion research. As illustrated in \Cref{fig:nudenet_vs_vlm_positionmap}, NudeNet can conflate socially acceptable exposure (e.g., swimwear or athletic attire) with harmful content, yielding high-confidence unsafe predictions. In contrast, Qwen3-VL provides a finer-grained judgment by assigning these cases to \emph{Safe} or \emph{Partial}, which better matches human evaluation and reduces incentives for over-suppression.

\section{Training Details of the Latent Classifier}
\label{appendix:classifier_training}

To ensure robust monitoring of latent states across all timesteps, we train a time-conditioned 4-class classifier designed to operate directly on the latent space. The classifier is trained on a balanced dataset of 6,000 images, with 1,500 samples per class.

\subsection{Dataset composition and class definitions}
The training data is synthesized using a pipeline that combines Large Language Model (LLM) prompting with Stable Diffusion v1.4 generation. The four classes are defined as follows:
\begin{itemize}
    \item \textbf{Benign (Class 0)}: To mitigate out-of-distribution issues and prevent the model from over-suppressing non-human concepts, we utilize ImageNet class labels as prompts.
    \item \textbf{Safe/Person (Class 1)}: This class consists of human subjects in socially acceptable attire.
    \item \textbf{Harmful/Nude (Class 2)}: This class includes explicit nudity and targeted harmful concepts that the model aims to suppress.
    \item \textbf{Color Artifacts Safe/Person (Class 3)}: This class represents semantic collapse. To improve structural stability, we include images of clothed people affected by color artifacts.
\end{itemize}

\subsection{Modeling guidance-induced color artifacts}
High guidance scales in safety frameworks often result in visual artifacts before total semantic collapse occurs. To proactively detect these failure modes, we simulate typical guidance-induced artifacts for Class 3 using the following corruption methods:

\begin{enumerate}
    \item \textbf{Local Color Corruption}: We inject high-intensity color noise into random patches within the latent-derived images to mimic localized pixel corruption caused by guidance conflicts.
    \item \textbf{Saturation and Hue Distortion}: We apply extreme saturation boosts (up to 4 times) and significant hue shifts in localized circular regions, simulating the color bleeding effect often seen in over-guided samples.
    \item \textbf{Neon Blotches}: We simulate neon-colored anomalies using Gaussian-masked blending with vibrant hues such as hot pink and neon green. These blotches replicate the characteristic artifacts generated by diffusion models when navigating aggressive safety boundaries.
\end{enumerate}

\end{document}